\documentclass[letterpaper, 10 pt, conference]{ieeeconf}  

\IEEEoverridecommandlockouts                              

\usepackage{graphicx} 
\usepackage{amsmath} 
\usepackage{amssymb}  
\IEEEoverridecommandlockouts 
\title{\LARGE \bf
SPIBOT: A Drone-Tethered Mobile Gripper for Robust Aerial Object Retrieval in Dynamic Environments
\thanks{© 2025 IEEE. Personal use is permitted. Permission from IEEE must be obtained for all other uses, including reprinting/republishing this material for advertising or promotional purposes, collecting new collective works, or reselling and redistributing to servers or lists.}
}

\author{Gyuree Kang$^{1}$, Ozan Güneş$^{1}$, Seungwook Lee$^{1}$, Maulana Bisyir Azhari$^{1}$, and David Hyunchul Shim$^{1*}$
\thanks{Accepted in IEEE International Conference on Robotics and Automation (ICRA) 2025}
\thanks{$^{1}$Gyuree Kang, Ozan Güneş, Seungwook Lee, Maulana Bisyir Azhari, and David Hyunchul Shim are with the Unmanned Systems Research Group, Department of Electrical Engineering, Korea Advanced Institute of Science and Technology (KAIST), Yuseong-gu, Daejeon 34141, Republic of Korea.
{\tt\small \{fingb20, ozan.guenes, seungwook1024, mbazhari, geninfty\}@kaist.ac.kr}}%
}

\begin{document}

\maketitle
\thispagestyle{empty}
\pagestyle{empty}

\begin{abstract}
In real-world field operations, aerial grasping systems face significant challenges in dynamic environments due to strong winds, shifting surfaces, and the need to handle heavy loads. Particularly when dealing with heavy objects, the powerful propellers of the drone can inadvertently blow the target object away as it approaches, making the task even more difficult. To address these challenges, we introduce SPIBOT, a novel drone-tethered mobile gripper system designed for robust and stable autonomous target retrieval.
SPIBOT operates via a tether, much like a spider, allowing the drone to maintain a safe distance from the target. To ensure both stable mobility and secure grasping capabilities, SPIBOT is equipped with six legs and sensors to estimate the robot's and mission's states. It is designed with a reduced volume and weight compared to other hexapod robots, allowing it to be easily stowed under the drone and reeled in as needed.
Designed for the 2024 MBZIRC Maritime Grand Challenge, SPIBOT is built to retrieve a 1kg target object in the highly dynamic conditions of the moving deck of a ship.
This system integrates a real-time action selection algorithm that dynamically adjusts the robot's actions based on proximity to the mission goal and environmental conditions, enabling rapid and robust mission execution. Experimental results across various terrains, including a pontoon on a lake, a grass field, and rubber mats on coastal sand, demonstrate SPIBOT's ability to efficiently and reliably retrieve targets.
SPIBOT swiftly converges on the target and completes its mission, even when dealing with irregular initial states and noisy information introduced by the drone.
\end{abstract}

\section{INTRODUCTION}
Aerial grasping using drones is a highly promising research area within field robotics, offering the flexibility to operate in dynamic and challenging environments often inaccessible to ground robots, such as rugged or shifting terrains and maritime settings. Drones in particular are favored for aerial grasping due to their precise maneuverability. However, aerial grasping with drones faces significant challenges when exposed to strong winds and dynamic conditions that complicate the process. Additionally, lifting substantial weights requires powerful propulsion systems, which can generate intense drafts that may displace or blow away target objects, particularly in critical environments like maritime settings, where a failed grasping attempt can lead to mission failure with no chance of recovery. These challenges underscore the need for a highly stable and robust aerial grasping system capable of reliable operation in harsh conditions.

\begin{figure}[t]
     \centering
         \includegraphics[width=0.48\textwidth]{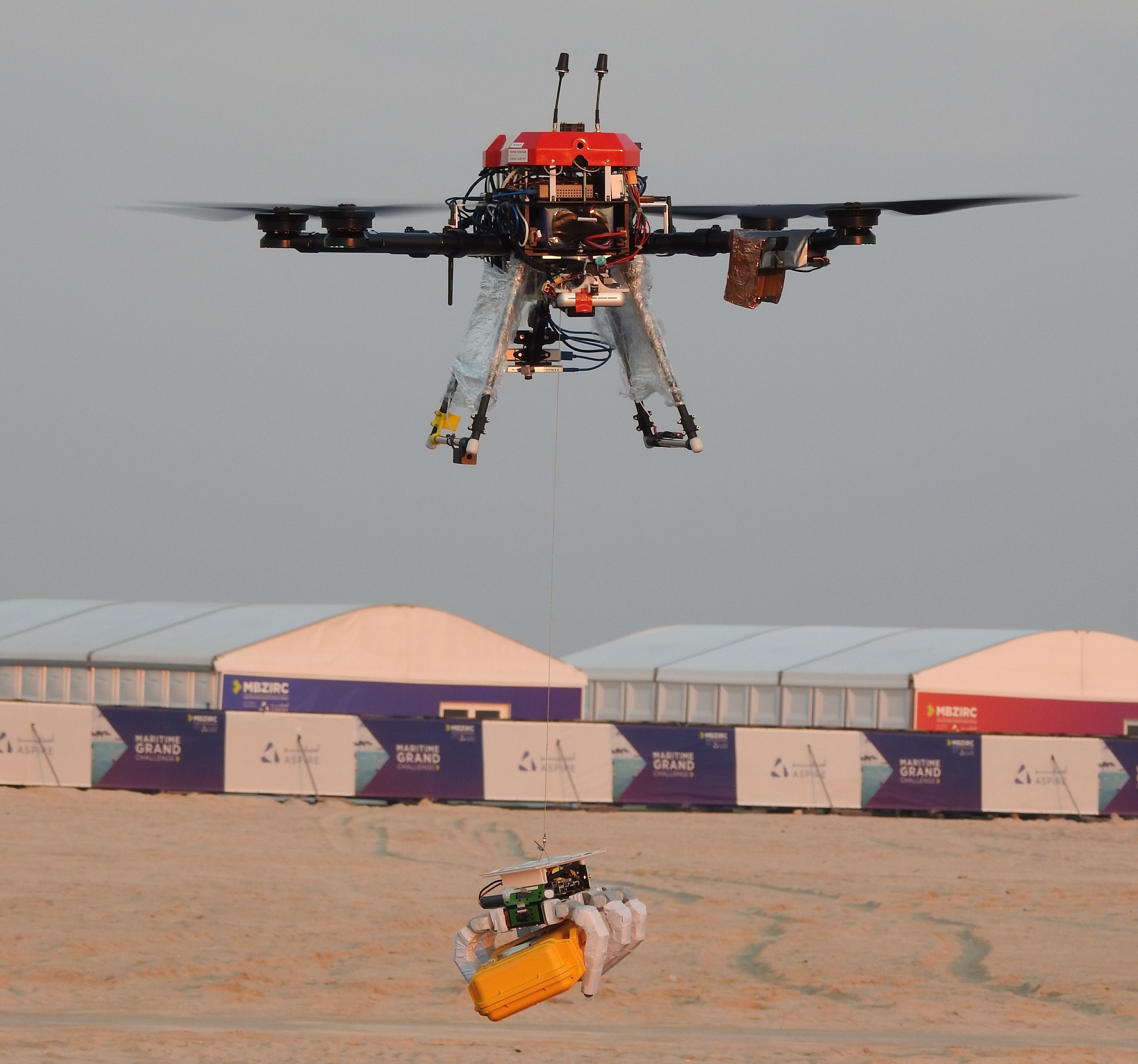}
     \caption[]{Drone retrieving the target box using SPIBOT}
     \label{fig:intro_image}
\end{figure}

In response to these challenges, this paper introduces SPIBOT (Fig. \ref{fig:intro_image}), a drone-tethered mobile gripper designed to address the problem of grasping objects in dynamic outdoor environments, such as the deck of a vessel in maritime settings. SPIBOT is designed for handling heavy and bulky objects across diverse environments by landing the robot via a tether—similar to a spider—rather than requiring the drone to directly reach the target. The system integrates a real-time action selection algorithm that adapts to dynamic conditions, ensuring robust and reliable performance in aerial grasping tasks, and includes retry mechanisms to prevent failures when instability is detected.

To summarise, the contributions are:
\begin{itemize}
\item \textbf{Introducing a drone-tethered mobile gripper system} that allows safe and stable target retrieval in challenging environments by maintaining a safe distance from the target.
\item \textbf{A real-time action selection algorithm} that dynamically adjusts the robot's actions based on proximity and environmental conditions, ensuring robust performance in dynamic scenarios.
\item \textbf{Real-world validation in various environments}, demonstrating the system's reliability and adaptability in complex, dynamic settings.
\end{itemize}

\section{RELATED WORKS}
For aerial grasping with drones, grippers are a popular choice due to their lightweight nature, which allows for efficient flight dynamics. However, grippers are typically limited by their grasping angle and struggle with handling large objects, making them less versatile when dealing with items of varying sizes and shapes  \cite{meng2022aerial}.
\cite{cheung2024modular} propose a modular pneumatic soft gripper designed for aerial grasping and landing, which is capable of handling objects of various shapes and sizes.
\cite{liu2020adaptive} introduce a novel suction cup-based aerial manipulation system that enhances tolerance to flight control errors during perching and grasping tasks. Their dual-cup design—featuring an inner soft cup and an outer firm cup—corrects angular errors, provides strong adhesion, and reduces the need for precise control, outperforming conventional suction cups in challenging conditions. Additionally, \cite{yadav2023integrated} combine a dual-stable gripper with adaptive control to achieve stable aerial grasping in various environments.
\cite{fishman2021dynamic} introduces a soft drone prototype that replaces traditional rigid landing gears with a soft tendon-actuated gripper, enabling dynamic and aggressive grasping. The system combines advanced quadrotor control algorithms with soft robotics models, allowing the drone to successfully grasp objects of unknown shapes. Experimental results demonstrate a high success rate, highlighting the effectiveness of the soft drone in scenarios where rigid systems typically fail.
\cite{hingston2020reconfigurable} presents a lightweight, dual-function adaptive landing gear and gripper prototype designed for multirotor UAVs, enabling them to land on slanted surfaces and moving platforms, as well as handle and transport various objects. They demonstrate the prototype's effectiveness in providing both adaptive landing and object manipulation capabilities, meeting the requirements for versatility and lightweight construction.

Angular Joint Arm Grippers offer greater flexibility in grasping angles, allowing for a broader range of motion. However, they are generally unable to grasp large objects and tend to be relatively heavy. \cite{lin2019autonomous} propose a vision-based grasping system using a monocular camera and a 3-DOF robotic arm. The system estimates the region of interest (ROI) and tracks objects using a correlation filter-based classifier, while a Support Vector Regression (SVR) based detector improves grasping success rates. 
Similarly, \cite{lin2018toward} present a rotation-aware grasping system using a 3-DOF robotic arm, employing a rotation-squeeze detection algorithm to detect target positions and angles in real-time. \cite{chen2019aerial} introduce a lightweight manipulator for aerial grasping, optimized for multiple objectives, including gripper posture tracking, joint limit avoidance, and maintaining the camera's field of view (FOV), all within a hierarchical control framework.

Meanwhile, multi-agent methods can grasp large objects of various shapes; however, they are difficult to control when dealing with moving objects \cite{meng2022aerial}. \cite{zhao2023versatile} introduces DRAGON, an articulated aerial robot with vectorable rotor units embedded in each link, enabling stable aerial manipulation and grasping. The key innovation lies in using a two-degree-of-freedom rotor vectoring apparatus and a comprehensive flight control method that suppresses oscillations and integrates thrust with external and internal wrenches for object handling.
\section{DESIGNS AND METHODS}
\subsection{Drone-Tethered Robot Configuration}

\begin{figure}[t]
     \centering
         \includegraphics[width=0.48\textwidth]{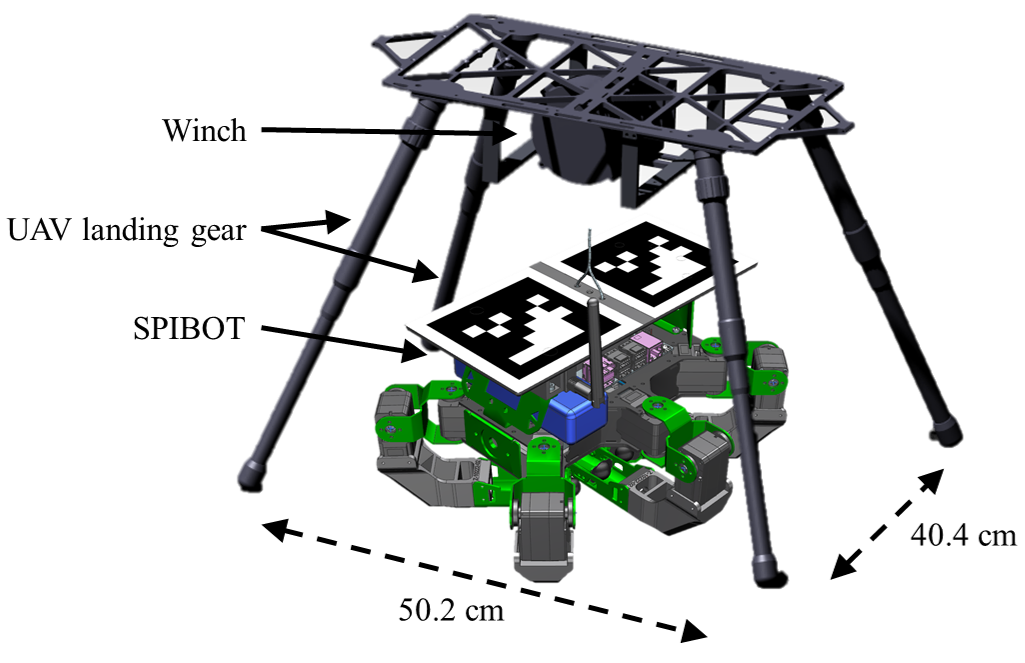}
     \caption[]{SPIBOT is fitted beneath the drone, with a winch positioned above it.}
     \label{fig:spibot_open}
\end{figure}

\begin{figure*}[t]
     \centering
         \includegraphics[width=1.0\textwidth]{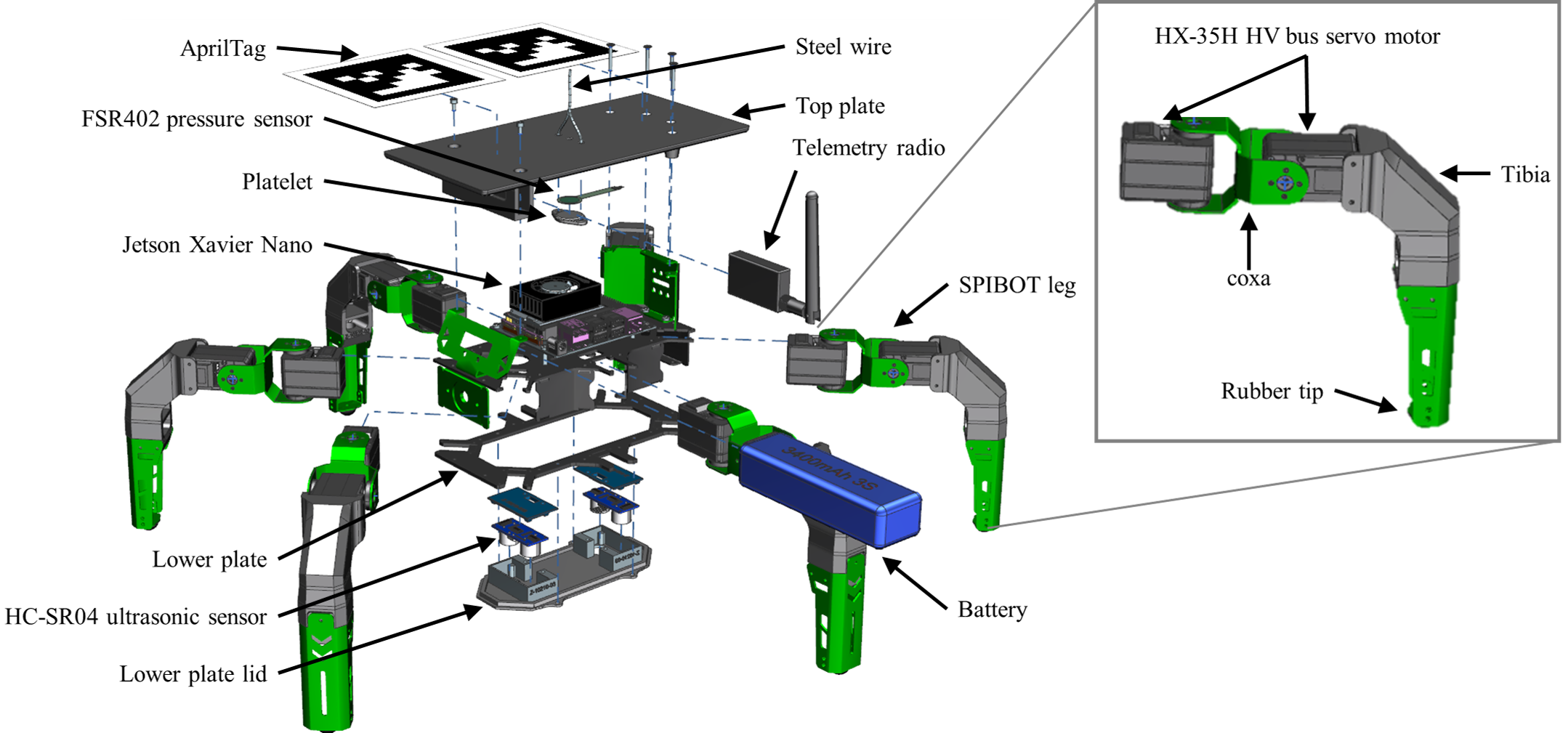}
     \caption[]{SPIBOT exploded view: The main body box construction, bearing the operation loads, accommodates the attachments for detection (top plate and AprilTags), sensing (platelet - pressure sensor and ultrasonic sensors), communication (telemetry radio) and gait. The red arrow denotes the connection to the carrier UAV through a steel wire.}
     \label{fig:spibot_exp}
\end{figure*}

To retrieve objects stably from the ground, the robot requires two key capabilities: mobility and secure grasping. We chose a hexapod robot for this purpose, as it effectively combines both. With multiple contact points, the hexapod can move stably across various terrains, maintaining balance even if some legs are impaired. This makes it highly mobile on dynamic and diverse surfaces. Additionally, its six legs can function as a versatile gripper, allowing the robot to grasp objects securely regardless of the target's shape or the gripping posture.

With set propeller size and therefore UAV dimensions, the primary development objectives for the hexapod robot as a mobile gripper are weight reduction and volume minimization. The requirement for low gripper weight is obvious as the gripper increases the UAV's operating empty weight, therefore directly influencing the maximum transportable payload. The volume minimization is necessary for the mobile gripper to fit beneath the UAV in loaded and unloaded condition. 
As a baseline we used a commercially available platform, namely the Hiwonder JetHexa, and modified it according to our needs for the aerial grasping task as illustrated in Fig. \ref{fig:spibot_open}.

To meet our requirements we used off-the-shelf motors, power electronics, a high-capacity battery as well as a lightweight frame made of light metal alloy parts and 3D printed Onyx and PLA components. We opted for a configuration with two motors per leg, resulting in a total of 12 Hiwonder HX-35H HV servo motors. These motors connect the coxa to the main body and to the tibia around the platform's circumference as in Fig. \ref{fig:spibot_exp}. 
The tibia link is designed as a 3D-printed Onyx and PLA component fixed at a 90 degrees angle with respect to the prior joint — a design choice that sacrifices some mobility for weight savings compared to the typical hexapod design, which uses three motors per leg.

For mission execution there are two HC-SR04 ultrasonic sensors inside the box construction as well as a FSR402 analog pressure sensor mounted within the power flow from the winch to the robot platform. Lastly the top plate, bearing two AprilTags, is attached to the platforms upside with two mounting brackets serving as detection target and as the attachment point for the steel wire connecting the robot to the UAV’s wire winch, which pull the robot with Robotis Mx-64AR servo motor.

Our carrier UAVs landing gear configuration is made up of four struts forming a 404mm x 502mm rectangle on the bottom which tapers prismatically towards the top. This space must accommodate the steel wire winch and the SPIBOT. The chosen SPIBOT design assumes bounding box volumes in open and closed configuration of 
\begin{center}(Length:Width:Height)$_{closed}$ = (300 x 268 x 175) mm 
\end{center}
\begin{center}
(Length:Width:Height)$_{open}$ = (390 x 410 x 225) mm 
\end{center}
leaving enough margin to all sides.

The pressure sensor is attached on a small platelet which in turn is connected to the tether. Upon reeling of the wire the platelet is pressed against the top plate generating a pressure sensor reading. This reading is being used to verify whether the robot is airborne or landed and, if airborne, whether a (specific) object is securely grasped and retrieved. 

Furthermore ultrasonic sensors are used to determine whether the target is positioned beneath the robot. In SPIBOT, two ultrasonic sensors are mounted, one on the front and one on the rear side of the robot, to ensure that the target is stably centered under the robot. This configuration allows the robot to grasp the target securely with all six legs. Additionally, the system is designed so that even if one of the two sensors fails, the remaining sensor can still detect the target. In the event of a pressure sensor failure, the robot will assess the stability of the grip either after landing or while airborne. This design enhances the robustness of the system.

The IMU sensor is used to measure the robot's angle, specifically detecting the tilt by measuring how much the robot has inclined relative to the gravity vector. This helps determine the instability of the robot. If the robot enters an unstable state, the system automatically sends a retry request to restart the mission. This mechanism prevents the robot from failing the mission due to an abnormal posture, ensuring that the robot can perform tasks stably.

The Nvidia Jetson Xavier Nano is used to manage tasks such as communication with the UAV via a 915MHz mRo SiK Telemetry Radio, sensor data processing, low-level gait control, grasping motion control, and running the robot's action selection algorithm. Power for the motors, sensors, and onboard computer is provided by an 11.1V, 3400mAh 3S1P battery.

For controlling the robot, we use a tripod gait, lifting three of its six legs simultaneously during locomotion. This gait strikes a balance between stability and speed, making it suitable for reliable and steady mission execution. However, with two degrees of freedom per leg, the robot is limited to forward and rotational movement, similar to differential wheel control, as movement in parallel directions is not possible.

\subsection{Robust Robot Action Selection Algorithm in Dynamic Environment}
\begin{figure*}[t]
     \centering
         \includegraphics[width=1.0\textwidth]{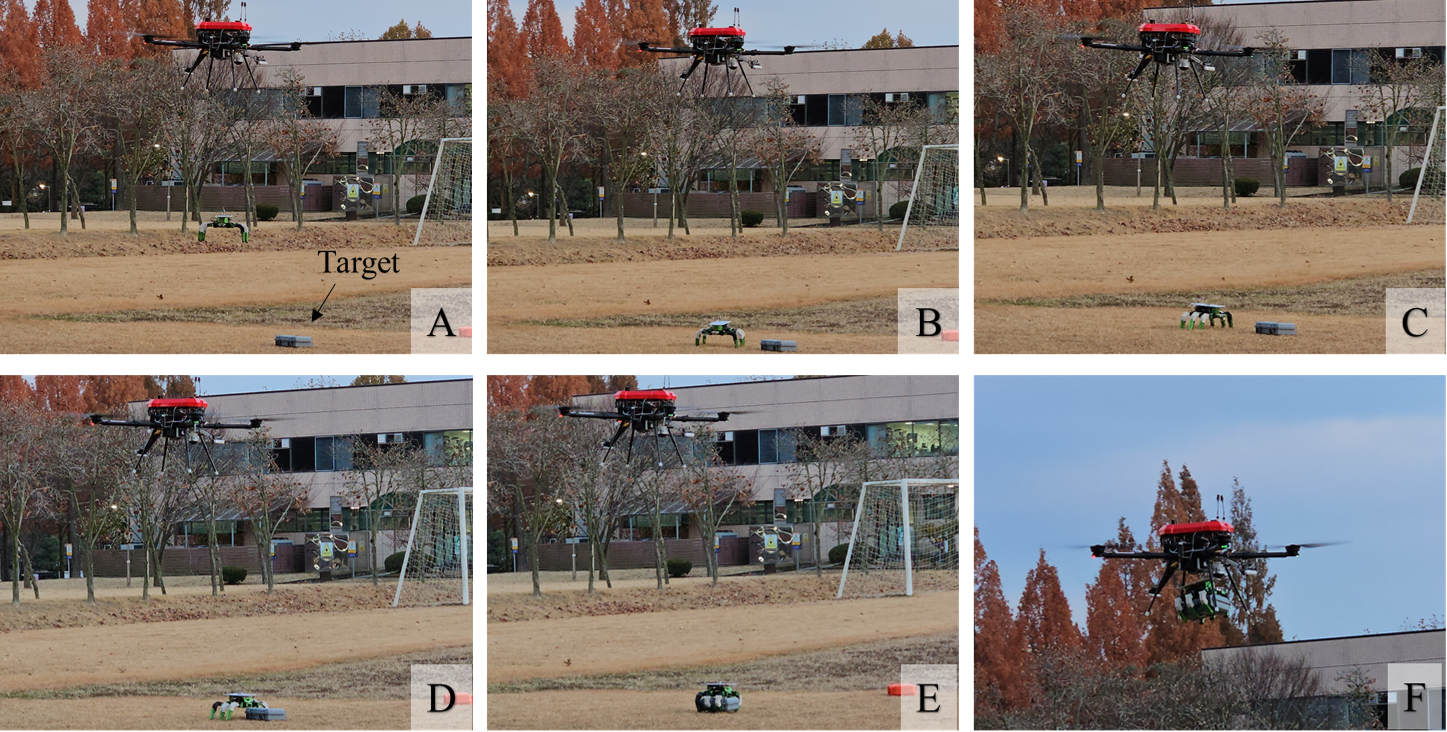}
     \caption[]{The drone-tethered mobile gripper retrieves a target object on the ground. Each image shows an action: (A) Stabilizing, (B) Aligning, (C) Approaching, (D) Forward Approach, (E) Grasping, and (F) Retrieving.}
     \label{fig:scenarios}
\end{figure*}
Aerial target retrieval, though seemingly straightforward, presents significant challenges due to factors such as wind and irregular target surfaces, which can lead to mission failures. Consequently, predefined sequence-based mission planners, typically used for simpler tasks, are inadequate in this context. To address these challenges, we developed a real-time action selection algorithm that ensures both rapid and robust performance by dynamically selecting actions based on the robot's current state and proximity to the mission goal.

Our approach utilizes two key concepts: goal proximity weight and action affordance \cite{gibson1977theory}, \cite{gibson2014ecological}.
Goal proximity weight $g$ prioritizes actions that advance the robot closer to the mission's objective, increasing the likelihood of selecting actions that directly contribute to goal attainment and thereby accelerating mission completion.
Action affordance $\alpha$ represents the feasibility of a specific action at given moment. By favoring actions with higher affordance, the robot enhances its ability to perform tasks robustly and reliably.
The Action selection module determines the optimal action by calculating the action priority, defined as the product of goal Pproximity weight and action affordance within the current state $S_t$.
\begin{equation}
    a_t = \arg\max_{i \in N} (g_i \alpha_i(S_t))
\end{equation}
This allows the robot to adaptively choose the most effective action in real-time, without relying on the rigid preconditions and effects used in traditional mission planners.

The robot actions include stabilizing, aligning, approaching, forward approaching, grasping, and retry.
During stabilizing in Fig. \ref{fig:scenarios}-A, the robot extends its legs to lower its center of gravity as it descends, preventing the tether from twisting and ensuring stability upon landing. 
In the aligning phase in Fig. \ref{fig:scenarios}-B, the robot orients itself toward the target, while the approaching phase in Fig. \ref{fig:scenarios}-C involves moving toward the target in the forward direction.
Once close enough, the robot performs a forward approach as in Fig. \ref{fig:scenarios}-D to position itself over the target, maintaining orientation even if the drone's view is obstructed for a few seconds.
When the ultrasonic sensor confirms the target, the robot executes the grasping action in Fig. \ref{fig:scenarios}-E, and the drone lifts both the robot and target during retrieving as in Fig. \ref{fig:scenarios}-F.
If an unstable state is detected or the robot fails to grasp the target, the retrying action initiates, signaling the drone to restart the mission.
The relative pose of the target is continuously provided by the drone, guiding the robot's actions in real-time.

The affordance for each action is calculated in real-time based on the mission and the robot's state. The state $S_t$ includes $s_a$, which reflects the heading error $\theta_t$ of the robot relative to the target, and $s_d$, which reflects the distance $d$ between the robot and the target.
\begin{equation}
    s_a = 1-\tanh(C_a {\theta} d / D )
\end{equation}
\begin{equation}
    s_d = 1-\tanh(C_d |d - T_d|)
\end{equation}
$T_d$ represents the desired distance for forward approaching, while $D$ denotes the maximum distance. The variable $s_u$ reflects the robot's stability, becoming 1 if the robot's tilt angle exceeds the threshold relative to the gravity vector. Additionally, $s_m$ indicates whether the grasping mission has started, $s_l$ indicates whether the robot has landed, and $s_o$ determines whether the robot is positioned above the target.
The affordance for each action $\alpha_i$ is calculated using the following equation, with the state weights presented in TABLE \ref{tab:affordance}.
\begin{equation}
    \alpha_i = \sum_{k \in \{m, l, a, d, o, u\}} w_k s_k
\end{equation}

\begin{table}[!htp]\centering
\caption{Weights for State Variables in Action Affordance}\label{tab:affordance}
\small
\begin{tabular}{|l|c|c|c|c|c|c|}\hline
Action & $w_m$ & $w_l$ & $w_a$ & $w_d$ & $w_o$ & $w_u$ \\ \hline
Stabilizing & 1 & 0 & 0 & 0 & 0 & 0 \\ \hline
Aligning & 1 & 1 & 0 & 0 & 0 & 0 \\ \hline
Approaching & 1 & 1 & 1 & 0 & 0 & 0 \\ \hline
Forward Approaching & 1 & 1 & 1 & 1 & 0 & 0 \\ \hline
Grasping & 1 & 1 & 0 & 0 & 1 & 0 \\ \hline
Retry & 1 & 0 & 0 & 0 & 0 & 1 \\ \hline
\end{tabular}
\end{table}
\section{EXPERIMENTAL RESULTS AND DISCUSSIONS}
\begin{figure*}[t]
     \centering
         \includegraphics[width=1.0\textwidth]{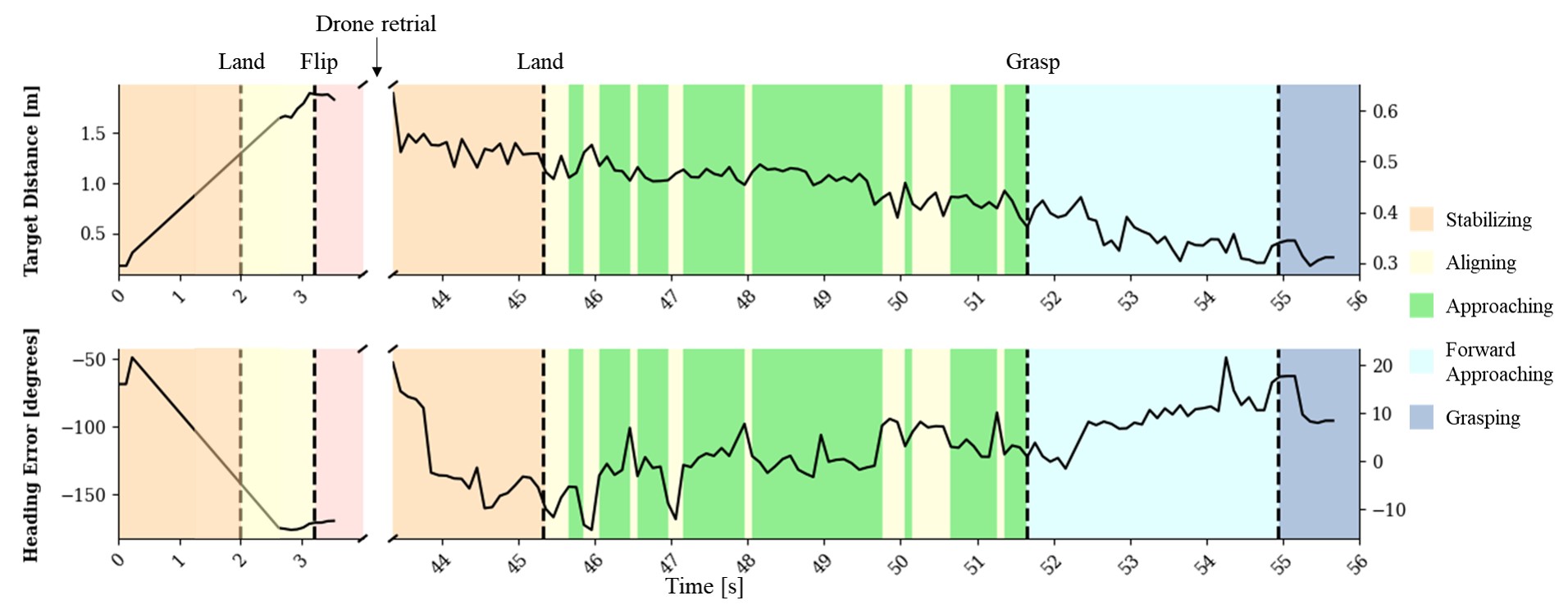}
     \caption[]{Target distance and heading difference relative to the robot. The orange, yellow, and green regions represent the timeline of the robot’s stabilizing, aligning, and approaching actions, respectively. The blue and dark blue regions indicate the forward approach and grasping actions. The red area represents the retreat action triggered when the robot became unstable.}
     \label{fig:result_plot1}
\end{figure*}
\begin{figure*}[t]
     \centering
         \includegraphics[width=1.0\textwidth]{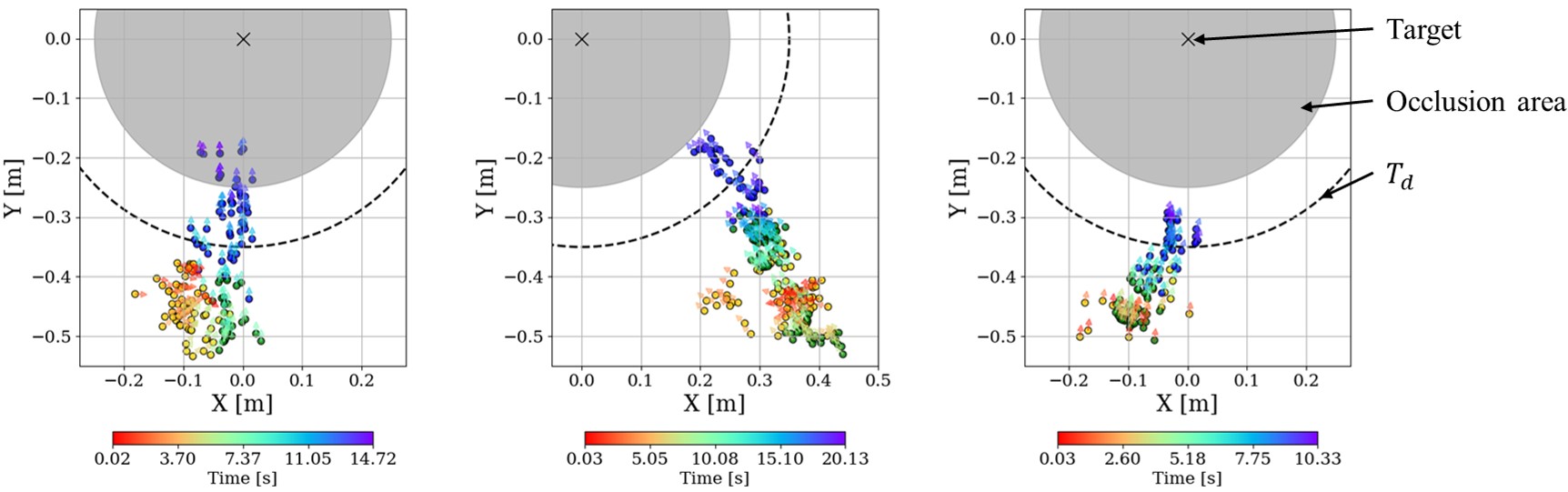}
     \caption[]{Robot trajectory relative to the target object. The arrows indicate the robot's heading direction, while the color gradient represents the progression of time.}
     \label{fig:result_plot2}
\end{figure*}

We tested the SPIBOT system in various environments, including a pontoon on a lake, on grass, and on a coastal testing site with irregular surface rubber mats installed on the sand. In the experiments, the drone operated autonomously, detecting the target object from the air, and hovering at a designated height of around 2.5 meters above the surface. The robot's mission started when it is landed at a point approximately 1 meter away from the target, with signal from the drone to start the mission.
The target object we used is a Pelican case with dimensions of 21.4 x 17.2 x 9.8 cm and an approximate weight of 1 kg.

Fig. \ref{fig:result_plot1} demonstrates the relative distance and heading error $\theta_t$ between the robot and the target, as measured by the drone. The orange, yellow, and green regions correspond to the stabilizing, aligning, and approaching actions, respectively. The action selection module dynamically adjusted the robot's behavior based on the state $S_t$, which includes $d_t$ and $\theta_t$, allowing the robot to quickly converge on the target.
The blue and dark blue regions represent the periods when the forward approach and grasping actions were executed. During the latter part of the forward approach and throughout the grasping phase, the robot occluded the target, causing the drone to lose visual tracking and making it unable to estimate the target's relative position. In response, two ultrasonic sensors located on the robot's body detected the target object underneath, initiating the grasping motion by setting $s_o$ to 1. Once the robot securely grasped the target, it signaled the drone to pull itself up with the target.
At the beginning of the figure, the robot experienced instability due to an uneven landing, resulting in a tilt. The action selection module responded by triggering the retreat action, setting $s_u$ to 1, allowing the drone to withdraw and restart the mission from the landing phase. This adaptive response enabled the drone to reattempt the mission, ultimately contributing to its success in the next trial.

Fig. \ref{fig:result_plot2} illustrates the robot's movement as observed by the drone during three different trials in varying environments. The arrows indicate the robot's heading direction, while the color gradient along the trajectory represents the progression of time. Despite the noise in the target position data from the drone, the robot consistently converged toward the target across all trials, demonstrating the effectiveness of the action selection algorithm in diverse environments. Notably, starting from the desired forward approach distance $T_d$, the robot’s heading aligned well with the target, allowing it to quickly move towards the target. The variations in the robot's initial trajectories reflect its adaptive response to different starting conditions in each trial. On average, it took the robot around 13 seconds to complete the mission on the ground.

\section{CONCLUSION}
We have presented SPIBOT, a drone-tethered mobile gripper inspired by the spider, designed for retrieving targets in harsh environments. SPIBOT's design and functionality addresses the challenges posed by dynamic and unpredictable environments, where other aerial grasping methods may struggle due to propeller-induced winds and shifting conditions. To ensure compatibility with drone operations, the robot's weight and volume were optimized for stable grasping and mobility, while maintaining the ability to perform in harsh environments. Additionally, the sensor configuration was carefully designed to enable reliable state estimation even under challenging conditions.
The integration of a real-time action selection algorithm allows SPIBOT to perform robustly, dynamically adapting to mission requirements and environmental changes. Our experiments, conducted in diverse settings such as lakes, grassy fields, and beaches, validate the system’s effectiveness, demonstrating its ability to complete missions swiftly and reliably. Although SPIBOT was initially developed for the 2024 MBZIRC Maritime Grand Challenge, its innovative approach to aerial object retrieval has significant potential to be expanded to address complex, real-world problems, advancing the capabilities of autonomous drones in various applications.

\bibliographystyle{IEEEtran}
\bibliography{spibot}


\end{document}